\documentclass[letterpaper]{article} 
\usepackage{aaai25}  
\usepackage{times}  
\usepackage{helvet}  
\usepackage{courier}  
\usepackage[hyphens]{url}  
\usepackage{graphicx} 
\usepackage{natbib}  
\usepackage{caption} 
\usepackage{algorithm}
\usepackage{algorithmic}

\usepackage{algorithm}
\usepackage{algorithmic}
\usepackage{multirow} 

\usepackage{newfloat}
\usepackage{listings}
\DeclareCaptionStyle{ruled}{labelfont=normalfont,labelsep=colon,strut=off} 
\floatstyle{ruled}
\newfloat{listing}{tb}{lst}{}
\floatname{listing}{Listing}
\title{Molly: Making Large Language Model Agents Solve Python Problem More Logically}
\author{
    Rui Xiao  \equalcontrib \thanks{*Corresponding author} \textsuperscript{\rm 1}, 
    Jiong Wang \equalcontrib \textsuperscript{\rm 1},
    Lu Han \textsuperscript{\rm 1},
     Na Zong \textsuperscript{\rm 1}
     Han Wu \textsuperscript{\rm 1}
}
\affiliations{
    \textsuperscript{\rm 1}Artificial Intelligence Research Institute, BEIDA JADE BIRD, Beijing, China\\

   \texttt{xiaorui@pku.edu.cn} \\
    \texttt{\{jiong.wang, lu.han, na.zong, han.wu\}@bdqn.cn}
}

\usepackage{bibentry}

\begin{document}

\maketitle

\begin{abstract}
Recently, applying large language models (LLMs) as teaching assists has attracted much attention as an integral part of intelligent education, particularly in computing courses. To reduce the gap between the LLMs and the computer programming education expert, fine-tuning and retrieval augmented generation (RAG) are the two mainstream methods in existing researches. However, fine-tuning for specific tasks is resource-intensive and may diminish the model`s generalization capabilities. RAG can perform well on reducing the illusion of LLMs, but the generation of irrelevant factual content during reasoning can cause significant confusion for learners. To address these problems, we introduce the Molly agent, focusing on solving the proposed problem encountered by learners when learning Python programming language. Our agent automatically parse the learners' questioning intent through a scenario-based interaction, enabling precise retrieval of relevant documents from the constructed knowledge base. At generation stage, the agent reflect on the generated responses to ensure that they not only align with factual content but also effectively answer the user's queries. Extensive experimentation on a constructed Chinese Python QA dataset shows the effectiveness of the Molly agent, indicating an enhancement in its performance for providing useful responses to Python questions.
\end{abstract}

%

\section{Introduction}
Large language models (LLMs) have exhibited extraordinary capabilities on language generation and understanding \cite{wang2022self, wang2024rat}. Therefore, faced with the increasing number of student questions on online Question and Answer (QA) platforms, exploring how to use LLMs as question and answer assistants has gained increasing attention \cite{hicke2023chata, dan2023educhat, agrawal2024cyberq}.

Particularly with rapidly growing number of learners in computing courses, such as Python, using LLMs as intelligent question-answering assistants shows the potential to revolutionize online education by providing comprehensive, and timely support to students. Therefore, exploring the use of LLMs models as programming teaching assistants has high practical value. 

\begin{figure}
    \centering
    \includegraphics[width=8.5cm,height=5.5cm]{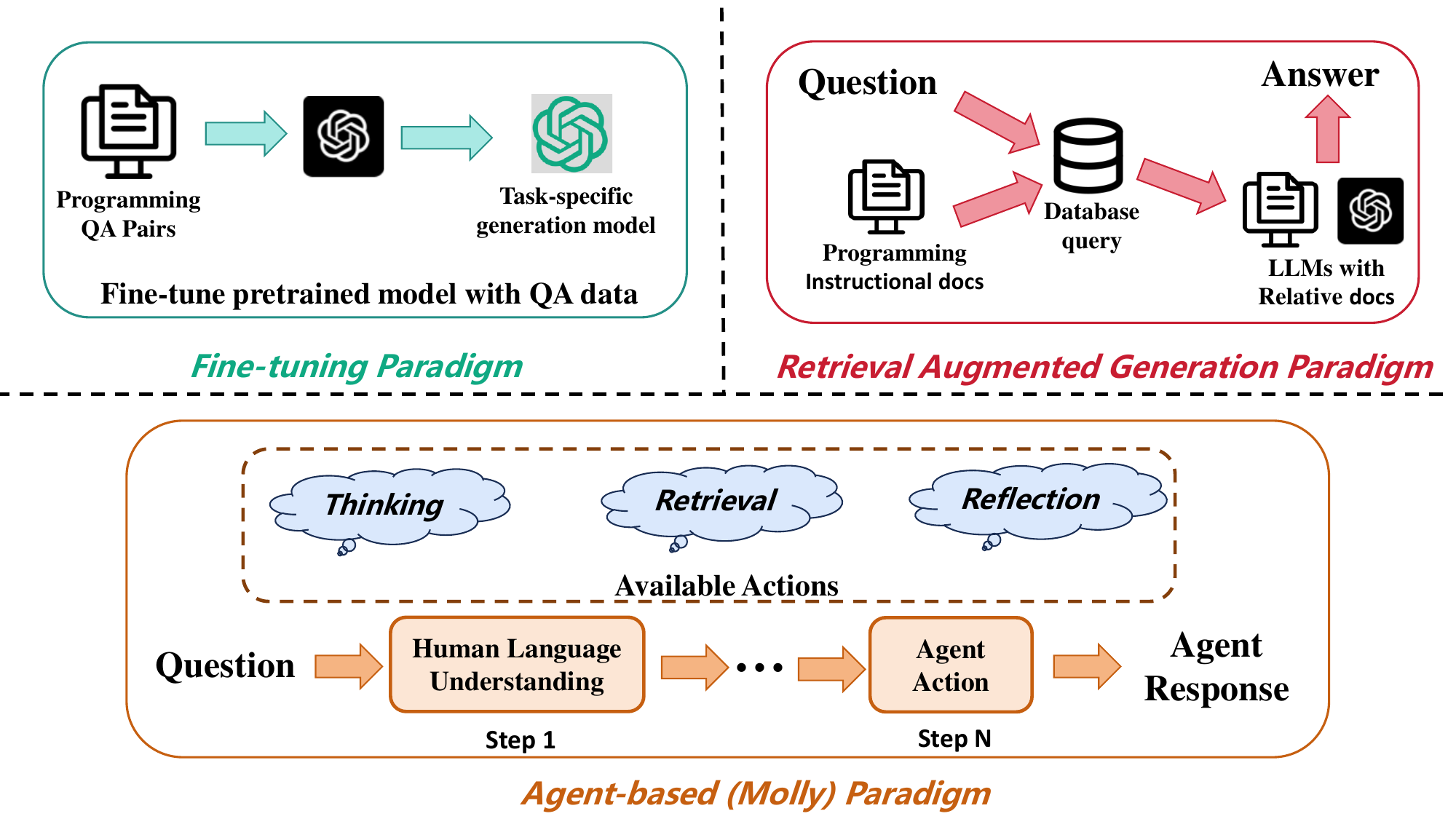}
    \caption{Molly-Agent compared with two mainstream programming QA employing LLMs.}
    \label{fig:intro}
\end{figure}

In actual application scenarios, despite the impressive accomplishments of LLMs, they often struggle to provide accurate responses to question that necessitate specialized expertise beyond their pre-training content, especially under programming teaching QA scenarios \cite{hicke2023chata}. To align the pre-trained knowledge of LLMs with the educational expert, existing methodologies \cite{hicke2023chata, baladon-etal-2023-retuyt, chang2024automatic} mainly focus on using supervised fine-tuning (SFT) or retrieval augmented generation (RAG), as shown in the upper sub-figures of Figure \ref{fig:intro}. Although the above methods can reduce the gap between the LLMs and the educational expert, there are still existing several challenges. 

One challenge is using SFT can significantly enhance the LLMs performance on specific datasets. But the knowledge in the field of education is updating, LLMs can not learn up-to-date knowledge under the alignment strategy \cite{chung2024scaling, dettmers2024qlora}. Therefore, in this application scenario, RAG seems more suitable to keep knowledge updated while reducing model hallucinations. However, when using RAG, the document content obtained during the retrieval phase cannot always be guaranteed to be associated with student questions \cite{guan2024mitigating, chen2024benchmarking}. This will lead to the LLMs generating factual incorrect answers, which is another challenge. \emph{So, how to improve the ability of LLMs as intelligent teaching assistants?}

To overcome the above-mentioned challenges, in this work, we introduce a LLM-agent framework called \textbf{Molly} to explore improving the ability of LLMs to answer learner Python questions in a more logical reasoning approach. Different from previous studies \cite{hicke2023chata, amalia2024implementasi}, which use LLM-related technologies to solve programming issues for English learners, we focus more on utilizing LLMs to address potential issues that Chinese Python learners may encounter during their learning process. 

In our agent, we combine learners' question understanding, RAG and iterative self-reflection. The process of our method as shown in Figure \ref{fig:intro}. We introduce a method based on role-playing scenarios to interactively detect learners' question intent. By leveraging LLM's understanding, this action can complete user queries through intent recognition, thereby enhancing the accuracy of knowledge retrieval. As mentioned above, the generated answers of RAG may be factual but unrelated to the learners' questions. Although the learner intent detection process has improved the accuracy of content retrieval and reduced the risk of generating irrelevant answers, in educational scenarios, ensuring the accuracy of teaching content alone may not always fulfill educational needs. Therefore, to further align the knowledge of LLMs with human education experts, we construct a newly structured QA dataset for teaching Chinese Python learners. Unlike previous studies \cite{dan2023educhat, hicke2023chata} that construct educational datasets to finetune LLMs, by creating more educational answers to help LLMs generated answers, we can align our agent with the teaching experience of education experts, ultimately enhancing the helpfulness of the generated answers.

For a complex generation task, solving it by using the self-reflection capabilities of LLMs has emerged as a promising solution \cite{zhou2022least, huang2024queryagent}. In our self-reflection module, by incorporating the guiding answers we have developed and making LLM learning the design principles of human education experts' answers, we improve the usefulness of the answers generated by LLMs. In addition, the agent goes through multiple rounds of reflection on its generated answers, assessing them for reasonableness, accuracy, and usefulness based on the retrieved answers and external context.

To the best of our knowledge, this is the first work exploring the use of LLM agents specifically designed for programming education.

Our contribution are summarized as follows:
\begin{itemize}
    \item We introduced Molly, a novel LLM agent framework that use LLMs for the educational Python QA task. The agent combines question intent detection and self-reflective answer generation to enhance the usefulness and helpfulness of generated answers. 

    \item We introduce a new Chinese Python QA dataset. By designing answer examples from multiple perspectives, this dataset can help guide LLMs towards providing teaching-oriented responses to learners' questions.

    \item We present a comprehensive evaluation of our method on questions from real learners, including both LLM-based and rubric-based human evaluations, on a range of models and baselines. 
\end{itemize}

\section{Related Work}
\subsection{Retrieval Augmented Generation}
Retrieval augmentation is a mainstream direction to connect LLMs to the external world. Facing frequently changing data, using fine-tuning to infuse knowledge into LLMs may lead catastrophic forgetting \cite{scialom2022fine, shao2023enhancing}. RAG addresses this issue by guiding the generation process of LLMs using the retrieved knowledge \cite{ovadia2023fine, hoshi2023ralle, chen2024benchmarking}. Without updating model parameters, RAG is an effective way for LLMs to interact with the updating real-world data \cite{toukmaji2024retrieval, li2024unigen}. By using the retrieved documents, LLMs can reduce the potential hallucinations and improve the relevance of generated content.

In this work, to further improve the accuracy of retrieval, the Molly agent first analyzes the learner's question using a scenario-based intent detection approach before conducting the search.

\subsection{AI-based QA assistants in education}
With an increasing number of learners turning to online QA platforms to get help with questions they encounter during their studies. AI chatbots and question-answering systems are becoming popular in education, providing support in answering what, why and how-to questions, administering quizzes and assessments, aiding faculty, and offering administrative services \cite{chen2021chatbot, mzwri2023internet, agrawal2024cyberq}. As more and more people enroll in programming courses, there exists a pressing need for innovative and scalable solutions to provide immediate and accurate answers to student programming queries. \citealt{hicke2023chata} combine RAG, SFT, and Direct Preference Optimization to develop a QA system specifically designed for programming context. However, existing studies have rarely focused on using AI to address educational issues in Python programming, especially for Chinese learners. \emph{The Molly agent, as a teaching assistant, is designed to help Chinese learners focus on solving Python programming problems.}

\subsection{Self-Reflection for LLM Agents}
As concerns persist about the accuracy and appropriateness of LLMs' generated content, self-reflection or self-correction has been proposed as a remedy to these issues \cite{ji2023towards, asai2023self}. Existing studies utilize prompt to rectify errors in the generated content \cite{chen2023teaching, pourreza2024din}. However, such intrinsic self-reflection, which is solely based on the inherent capabilities of LLMs without the crutch of external feedback, is hard to achieve significant improvement and is unreliable \cite{huang2023large}. Therefore, feedback-based self-correction method \cite{huang2024queryagent} can proactively identify when errors arise and provide tailored guidelines. For programming education scenarios, exploring the self-reflective generation process of LLMs is crucial, as both the accuracy of knowledge point explanations and the correctness of example codes need to be strictly ensured, while also ensuring that the generated content is helpful to learners.

In our work, the molly agent can self-reflection on the answers generated by LLMs from the perspectives of content rationality, code correctness, and answer usefulness, and continuously refine the results based on the answers retrieved from the knowledge base, enabling the generated results to better address learners' questions.

\begin{figure*}
    \centering
    \includegraphics[width=.85\linewidth]{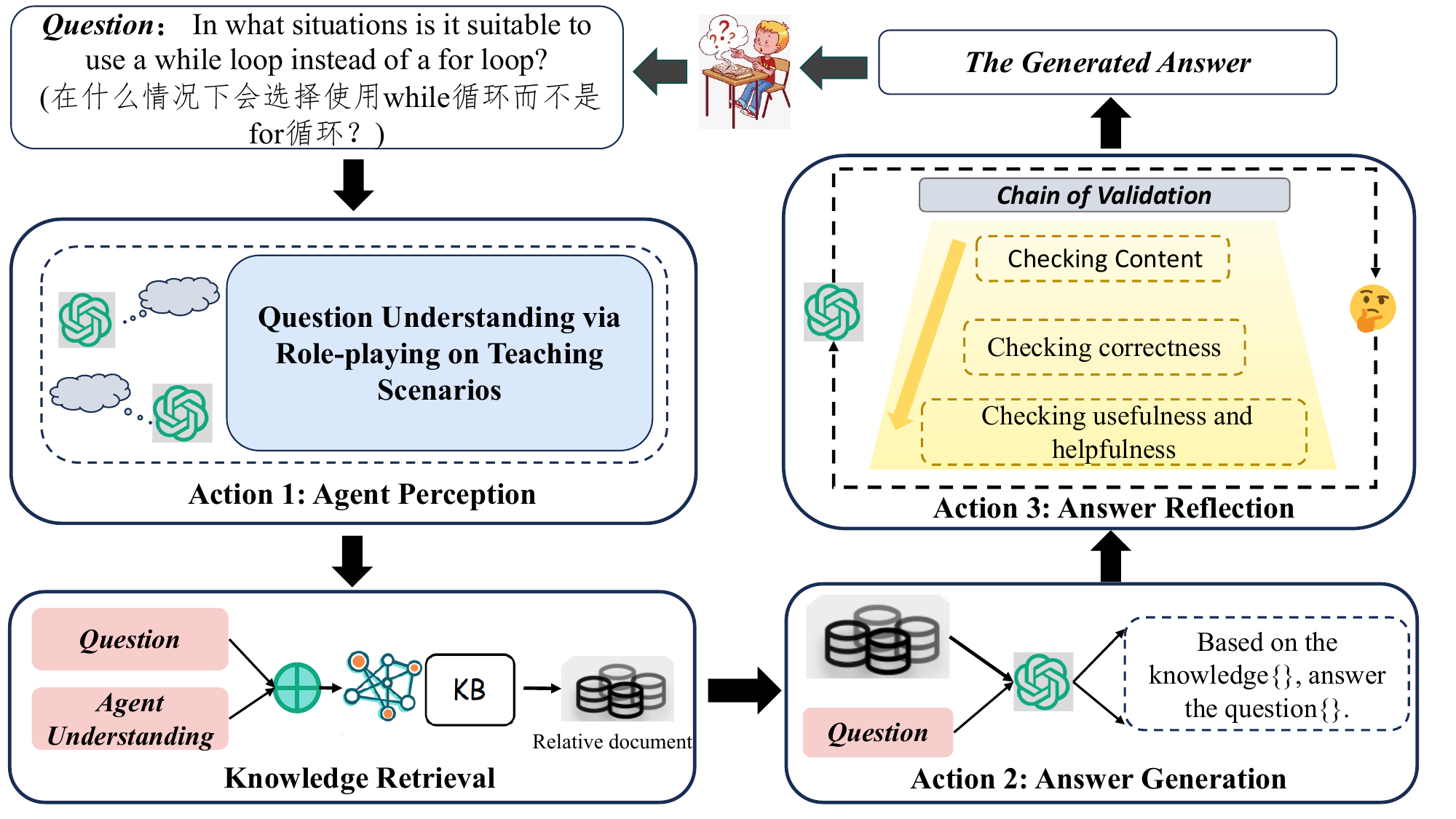}
    \caption{The framework of MollyAgent. Molly operates in three stages: (1) Agent Perception. In this stage, we use large models to further clarify the user's questions and provide more relevant information for the learner's questions to improve retrieval accuracy. (2) Knowledge Retrieval and Answer Generation. At this stage, we combine the learner's questions with the LLM's understanding to retrieve relevant answers from the constructed structured knowledge base. We then use the LLM to generate appropriate answers based on the retrieved relevant answers. (3) In the final stage, we iteratively refine the answers generated in the previous phase using role-playing capabilities to ensure their accuracy and usefulness.}
    \label{fig:method}
\end{figure*}

\section{Data Collection}

In existing related works, datasets for diverse Python programming learners are still scarce. To address this challenge, this paper introduces a specialized Chinese single-turn question-and-answer (Q\&A) dataset for Python learners, especially for beginners. This dataset provides 5,960 questions along with their corresponding structured, specialized answers.

\textbf{Question Organization.}
To ensure that the questions from real-world questioning patterns of Python learners, we gathered 16,247 Python-related questions from offline and online collections. These questions cover all knowledge points of Python course, ranging from \emph{basic concepts, Python syntax to complex programming issues, and from understanding data structures to the application of algorithms and other aspects}. Subsequently, we categorized and organized these questions. To ensure the clarity and representativeness of questions, the selection process was carried out by experienced Python educators. This process involved classifying, filtering, and restructuring the questions to ensure that each one has a clear expression, specific learner background, and a definitive question description. We excluded repetitive, ambiguous, or questions not directly related to Python learning to ensure that the final selection of questions is both high-quality and highly relevant. Each question in the dataset is associated with an ID, corresponding knowledge point type and an educational answer.

\textbf{Answer Design Principles.}
To ensure the quality of the designed answers, we implemented a rigorous quality control process. Considering the need for profound domain knowledge in Python programming and substantial teaching experience for the creation of the dataset, we selectively recruited 8 professional instructors specializing in Python programming as our annotation experts. Prior to commencing the official annotation process, these experts underwent thorough training. This training was designed to ensure they fully grasped our specific annotation requirements.

Throughout the annotation process, we implemented a dual-verification mechanism to guarantee the accuracy and dependability of the results. Every annotation underwent two separate rounds of verification post-submission. Initially, an annotation expert conducted the first round, meticulously examining each submission, and suggesting revisions for any errors or ambiguities identified. After implementing these corrections, a second expert embarked on the subsequent verification phase. This stage involved a thorough review of annotations that had successfully passed the initial round, to ensure compliance with our stringent standards. Annotations that were validated as accurate in this second phase were then officially approved.

\textbf{Answer Quality Verification.}
To ensure the consistency of annotations, after annotating a certain amount of data, we randomly sampled 1,000 QA pair for consistency verification. Two annotation experts independently scored these annotations for consistency, calculating the overall quality of the answers. The scores were then used to compute the kappa value for annotation consistency. The final score was 0.90 (The score closer to 1 indicates higher consistency), demonstrating that our annotations are highly consistent, thereby ensuring the objectivity of the answers to the greatest extent possible. Table \ref{tab: data} shows the overall statistical data of the annotated dataset. 

\begin{table}
    \centering

    \small
    \begin{tabular}{c|c}
    \hline
    \multirow{2}{*}{\textbf{Statistic type}} & \\
        &   \textbf{Question} \\
         \hline
         Number of dialogues &  5,960\\
        \hline
         Longest question length & 54.5\\
         Shortest question length & 6\\
         Average question length & 18.15\\
        \hline
            &  \textbf{Answer}\\
        \hline
         Max. \# tokens per answer & 1306.5 \\
         Min. \# tokens per answer & 247\\
         Avg. \# tokens per answer & 508.08\\
         \hline
         Number of answers containing code  & 3516\\
    \hline
    \end{tabular}
    \caption{Overall statistics of the structured dataset.}
    \label{tab: data}
\end{table}

\section{Molly}
The overall frame of Molly as shown in Figure \ref{fig:method}. Molly operates in three stages. First is Agent Perception, where large models are used to clarify the user's questions and provide more relevant information to enhance retrieval accuracy. The second stage involves Knowledge Retrieval and Answer Generation. Here, the learner's questions are combined with the LLM's understanding to retrieve pertinent answers from a structured knowledge base, and the LLM generates suitable responses based on these retrieved answers. In the final stage, the answers from the previous phase are iteratively refined using role-playing capabilities to ensure their accuracy and usefulness, improving the teaching capability of the generated answer.

\subsection{Agent Perception}
Although large language models (LLMs) have demonstrated exceptional performance on tasks including question answering (QA), they can still generate factually inaccurate responses because their knowledge relies solely on parameter memory \cite{huang2024queryagent, zhang2024raft}. Moreover, it may be unrealistic to remember all (and constantly changing) world knowledge. To address this issue, retrieval-augmented LLMs have gained increasing attention by incorporating additional retrieval modules into LLMs, combining non-parametric knowledge \cite{jeong2024adaptive, lyu2024crud}. Specifically, these models access a knowledge base, which serves as a comprehensive repository of information across various disciplines and fields. This knowledge base is used to retrieve information relevant to a given input, which is then integrated into the LLMs, helping them maintain the accuracy and timeliness of world knowledge. 

In real-world applications, learners of programming languages often find it challenging to ask complex questions; the questions they pose are usually simpler and more straightforward. As a result, retrieving relevant information without sufficient context can be quite difficult. Since we have constructed a structured knowledge base, the retrieval phase involves matching against questions in the knowledge base rather than the documents themselves. As shown in Table 1, the token length of questions in the dataset indicates that the relevant knowledge provided by the questions in the knowledge base is limited. Consequently, simply using the learner's query to retrieve information from the questions in the knowledge base often results in matches with similar but unrelated questions, which can affect the accuracy of answer retrieval.

Therefore, to provide learners' questions with sufficient context, we have introduced a user intent recognition mechanism within the proposed framework. We introduce a method based on role-playing scenarios to interactively detect learners' question intent. By leveraging LLM's understanding, this action can complete user queries through intent recognition. In the role-playing mechanism, we use one instance of a LLM to act as a student and another instance to act as a teacher. The teacher model is used to identify which knowledge points or perspectives can address the current problem, rather than providing a direct answer. The student model, which represents a knowledgeable learner with some experience in Python, evaluates whether the teacher model’s approach can effectively address the question from the perspective of the questioner. This process results in a note-style summary of the teacher model’s output, ensuring the content is concise and refined. 

\subsection{Knowlegde Retrieval and Answer Generation}
After obtaining the summary of the first stage, we combine the questions and answers as queries to match with the questions in the constructed knowledge base. We use cosine similarity to match relevant questions and find the corresponding structured answers. By selecting the top-3 most relevant questions based on the computed similarity scores, we obtain the answers for these questions. This allows us to generate preliminary results based on the retrieved answers.

\subsection{Answer Reflection}
\begin{figure}
    \centering
    \includegraphics[width=1\linewidth]{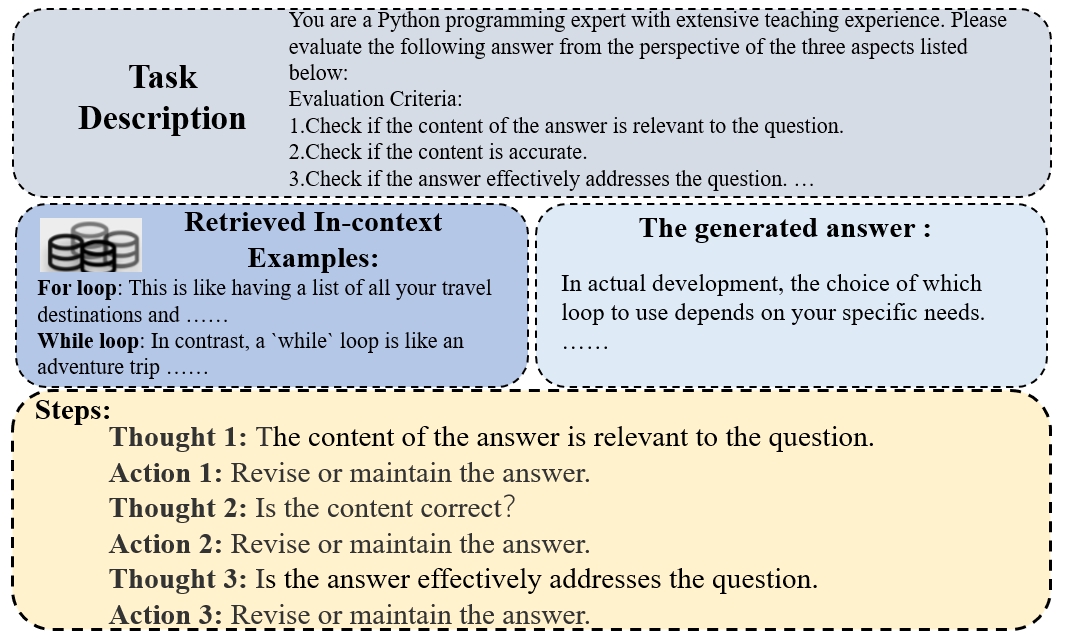}
    \caption{The Prompt Template of Answer Reflection Stage. We illustrate Molly’s prompt template during the final stage. Through phased observation and evaluation, answers generated in the second stage are refined using the self-reflective capabilities of a large model. This further optimization improves the model's ability to respond to learners' questions.}
    \label{fig:val}
\end{figure}

Recently, intelligent agent frameworks based on large model self-reflection mechanisms have received increasing attention. These frameworks are designed to handle situations where the initial large model's response cannot resolve the user query. The self-reflection mechanism sequentially reactivates AnyTool to re-answer \cite{toukmaji2024retrieval} or refine the response to the user query \cite{du2024anytool}. Notably, this mechanism can be applied repeatedly until a termination condition is met. 

In educational contexts, ensuring the accuracy of content generated by large models is crucial, as the correctness of the generated content directly impacts the learner's understanding of the knowledge. To further ensure the educational value and usefulness of the content generated by the large model, we propose an iterative feedback-based self-reflection mechanism to review and correct the answers generated in the second stage. The self-reflection process proposed in this paper is illustrated in Figure \ref{fig:val}. To ensure that the content generated during the self-correction process remains aligned with the knowledge and educational approach of human programming experts, the iterative correction process consistently references the human-annotated sample answers retrieved in the previous stage. This approach guarantees that the generated content maintains its educational value.

\section{Experiments}

\subsection{Experimental Setup}

\noindent\textbf{Baseline Models.} Since the proposed MollyAgent is model-agnostic, we can employ various LLMs in this framework. The used LLMs is as follows:

GPT-3.5-turbo\footnote{\url{https://platform.openai.com/docs/models/}}: Developed by OpenAI, GPT-3.5 is a language model capable of a wide range of natural language processing tasks such as translation, summarization, Q\&A, and text completion. It generates coherent responses relevant to given inputs.

GPT-4: An upgrade of GPT-3.5, GPT-4 not only maintains the powerful natural language processing abilities of its predecessor but also has significant improvements in model architecture, algorithm optimization, and data handling. It can more effectively understand complex text structures, contextual meanings, and user intentions, thus providing more accurate and rich responses.

ChatGLM \cite{glm2024chatglm}: An open-source bilingual model based on a general language model framework. ChatGLM has undergone specific optimizations including supervised fine-tuning, feedback-guided, and reinforcement learning combined with human feedback. We used ChatGLM-4 in our experiments.

For main experiments, we also adopt the RAG approach proposed by \citealt{hicke2023chata} as our baseline method. We extract text from PDFs about Python textbook. We converts PDFs into markdown files and is adept at capturing LaTeX math, tabular data, and code snippets found in the PDFs. To enhance retrieval efficiency and accommodate the limited context length of LLMs, we divide the source documents into chunks with a maximum length of 1,000 characters each and an overlap of 100 characters between adjacent chunks. To preserve paragraphs, sentences, and words, we use predefined delimiters—such as double newlines, single newlines, spaces, and empty strings—to split the document until the resulting chunks are appropriately sized. We use embedding-based retriever to obtain related answers from the knowledge base. We constructed the context using a union of the top-3 documents retrieved by the embedding based method.

\noindent\textbf{Experimental Details.} In our experiments, we carefully selected 334 Q\&A examples from the constructed dataset for experimentation. These examples cover common knowledge points of Python course and were chosen through random sampling to ensure the diversity and comprehensiveness, thereby providing a thorough evaluation of the performance of Molly.

\begin{table}
    \centering
    \large
    \begin{tabular}{c|ccc}
    \hline
         Metrics  &  AC &  EA & UF \\
         \hline
         Weights &  70\% &  10\%&  20\%\\
    \hline
    \end{tabular}
    \caption{Metric principle. AC (Answer Correctness) means to metric technical accuracy and rigor of answer. EA (Expressive Ability) means the logical organization and expressiveness of language. UF (Usefulness) means whether the content of the answer solves the problem. } 
    \label{tab:metrics}
\end{table}

\begin{table*}
    \centering
    \begin{tabular}{c|cccccc}
    \hline
        Method & AC & EA & UF & Overall Score & Code Accuracy (\%)\\
        \hline
        GPT-4 & 78.46 & 75.82  & 67.95  & 76.09 &  96.3 \\
        GPT-3.5-turbo & 72.03 & 75.29  & 67.22  & 70.63 & 92.3\\
        ChatGLM4 & 73.91 & 76.91 & 68.85  & 71.02 & 92.5\\
    \hline
        GPT-4 w/ RAG & 80.46 & 77.82 & 71.95  & 78.28 &  97.1 \\
        GPT-3.5-turbo w/ RAG & 75.91& 75.91 & 68.85  & 74.49 & 94.2\\
        ChatGLM4 w/ RAG & 77.11 & 77.51 & 68.95 & 75.52 & 94.7\\
    \hline
        GPT-4 w/ Molly & 86.32 & 85.42 & 77.65 & 84.49 & 96.6\\
        GPT-3.5-turbo w/ Molly & 82.00 & 86.50& 74.06 & 80.85 & 93.7\\
        ChatGLM4 w/ Molly & 83.63 & 85.50&  79.41&  82.97 & 94.3\\
    \hline
    \end{tabular}
    \caption{Python problem solving ability evaluation results of LLMs. Overall Score means the comprehensive satisfaction index, which is computed by Eq(1). AC means technical accuracy and right of answer, EA indicates expressive ability, UF indicates the answer can solve the problem. Code Accuracy means the accuracy rate of Python code generation.}
    \label{res1}
\end{table*}

\noindent\textbf{Evaluation Metrics.}
In our work, we employ human evaluations to assess the overall quality of answers generated by the MollyAgent and compared methods. To ensure that the generated answer conforms to our manually constructed annotation principles, we conduct a comprehensive scoring evaluation from three aspects: the correctness of answer explanations, the consistency of questions and answers, and the usefulness of answers, followed by calculating the overall score (the higher the score, the better the quality). In this process, we give the same prompts\footnote{To ensure fairness in our experiments, we maintained a consistent prompt for different LLMs. The prompt used in the experiment is: "You are a Python teacher, and I am a Python learner, please answer my question:". For the RAG-based method, the prompt we use is: “You are a Python teacher, and I am a Python learner. Please answer my question based on the retrieved relevant documents. Documents: {\emph{ \{ documents \}}}. Question: {\emph{ \{ question \}}} ”} and questions to different methods for generating responses, and finally, we use human evaluation to judge the generated results.

The specific quantitative metric design principles for the generated results are as follows:
\begin{enumerate}
    \item We quantify the above three aspects of answer generation into the following four evaluation dimensions: Technical accuracy and rigor of answer (Answer Correctness, AC), Logical organization and expressiveness of language (Expressive Ability, EA), Whether the content of the answer solves the problem (Usefulness, UF). 
    
    \item The final evaluation metric is a comprehensive satisfaction index: Excellent, Good, Average, Poor (out of 100; Excellent: 90-100; Good: 80-90; Average: 60-70; Poor: below 60). 
    
    \item For questions that require code generation, we score the accuracy of the generated code separately (1 for correct, 0 for incorrect), and finally report the accuracy rate of code generation.
\end{enumerate}

The weight distribution standards for different dimensions are shown in Table \ref{tab:metrics}.
And the calculation result of the final composite score is shown in Equation (1).

\begin{equation}
    \emph{overall} = 0.7 * AC + 0.1 * EA + 0.2 * UF 
\end{equation}

\subsection{Experiments Results}
Table \ref{res1} reports the differences between the methods discussed in this paper and the comparison baseline methods. We evaluated on three large language models and, through analyzing the data in the table, observed significant performance differences in Python problem-solving capabilities across different large language models and their respective enhancement frameworks. 

\noindent\textbf{Assessing the Python problem-solving capabilities of LLMs.} Firstly, we evaluated the Python problem-solving capabilities of large language models. The data from the table shows that GPT-4 performs exceptionally well, particularly in terms of answer accuracy, significantly outperforming other language models. However, in terms of expressive content, GPT-4 has improved slightly over GPT-3.5 and is noticeably inferior to ChatGLM4, indicating that Chinese large language models are more suitable for expressing answers to Chinese learning questions. In terms of code accuracy, GPT-4 continues to excel, demonstrating its strong code generation capabilities. 

\noindent\textbf{Comparison between RAG and Molly.} Next, we compared various enhancement methods for large language models. GPT-4 consistently demonstrated optimal performance across different scenarios. In contrast, the Retrieval-Augmented Generation (RAG) method, which relies on external knowledge bases to generate information, showed a notable improvement in answer accuracy. However, because the final output still depends on the language model itself, there are some limitations in its expressive capability.

When employing an iterative reflection mechanism within the agent framework, the language model adjusts its output dynamically. This approach resulted in clear advantages in both expressive ability and the overall usefulness of the answers provided. It is important to note that the retrieval methods we compared involved extracting relevant documents from textbooks. Variations in performance may be attributed to differences in the knowledge bases used.

Additionally, our method, which utilizes the designed dataset for the knowledge base, has a distinct advantage in incorporating human expert knowledge. In the subsequent sections, we will further analyze the impact of the constructed knowledge base.

Furthermore, evaluations of GPT-3.5-turbo and ChatGLM4 using both the RAG and Molly frameworks indicate that leveraging external knowledge bases enhances the accuracy of responses to Python-related questions. For additional details and case studies, please refer to the Supplementary Materials and Appendix A.1.

\begin{figure*}
        \centering
    \includegraphics[width=.9\linewidth]{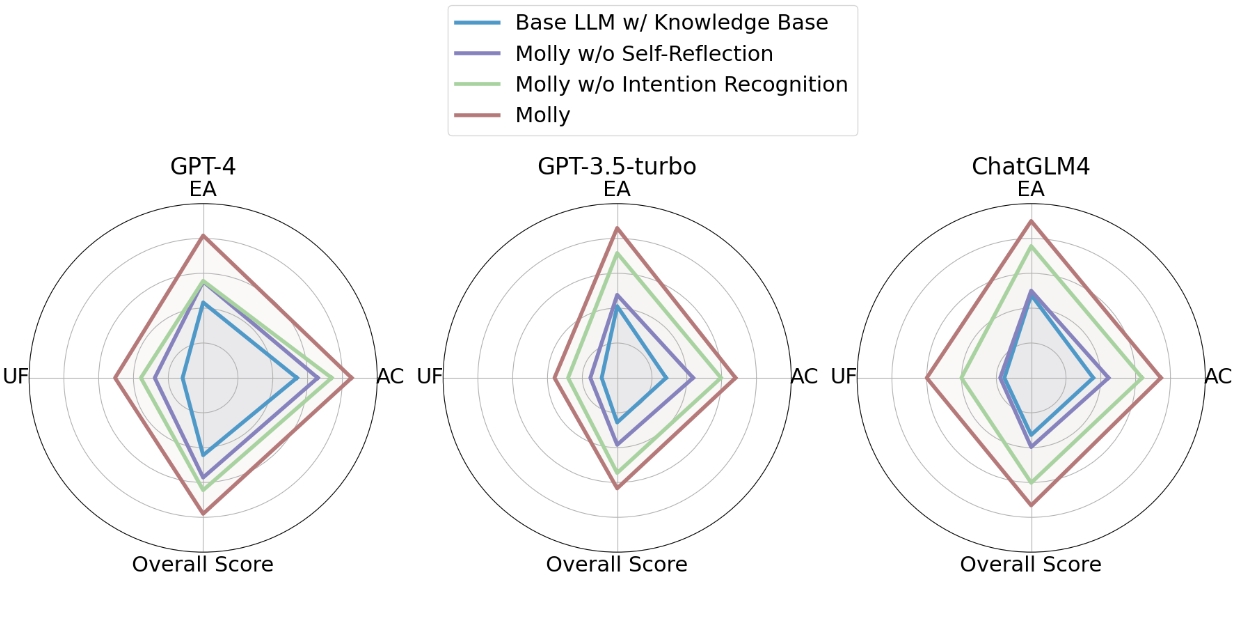}
    \caption{The effectiveness of the intention recognition based on role-playing and self-reflection.}
    \label{fig:aba}
\end{figure*}

\subsection{Ablation Study}
To verify the effectiveness of each module, we conducted ablation experiments, comparing the baseline LLM and Molly by sequentially removing the intent recognition module and the self-reflection module. The specific experimental results are shown in Figure \ref{fig:aba}. Compared to LLM, Molly w/o Intention Recognition showed no significant improvement, while Molly w/o Self-reflection demonstrated a more noticeable enhancement. This indicates that within the framework of an intelligent agent based on large language models, enhancing the output of the language model can significantly improve the ability to resolve Python-related issues for learners. However, when the intent recognition module and self-reflection are combined, the model's performance sees a further improvement. This suggests that incorporating intent recognition can help retrieve more relevant questions, enabling the model to obtain more targeted answers and thereby enhance the quality of the contextual examples provided.

\begin{figure}
    \centering
    \includegraphics[width=1\linewidth]{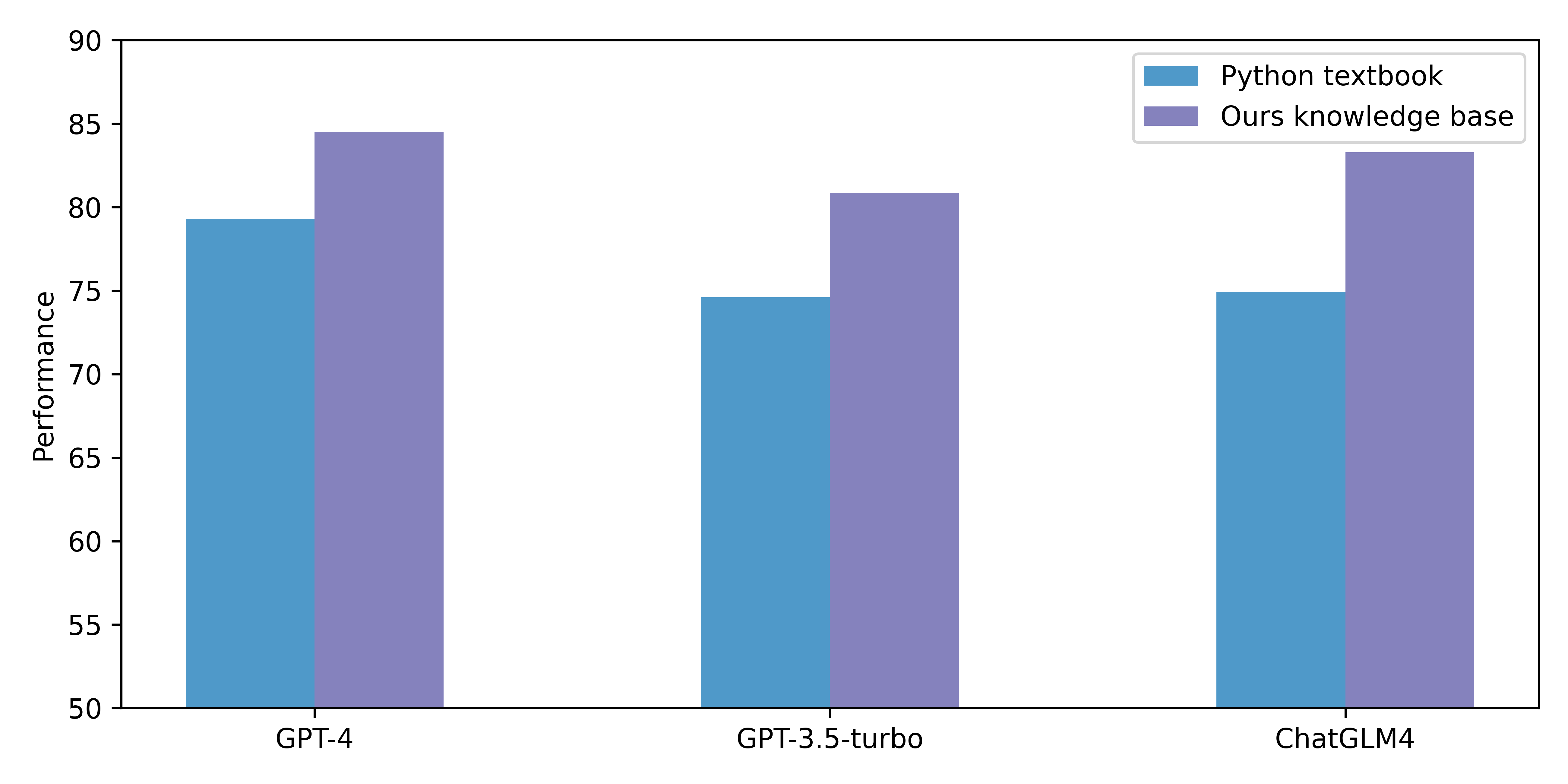}
    \caption{Comparative results of using different methods as knowledge bases.}
    \label{fig:com1}
\end{figure}

\subsection{The Effect of Knowledge Base}
We further compared the performance under different knowledge bases, with the results shown in Figure \ref{fig:com1}. We assessed the overall performance of various LLMs. The findings indicate that using our constructed dataset as a knowledge base helps enhance the pedagogical significance of large models in answering Python-related questions. Due to the educational nature of our knowledge base, it can be observed that when using our knowledge base to answer Python programming questions, ChatGLM's performance is nearly on par with GPT-4, demonstrating the value of the knowledge base we have constructed. 

\maketitle
\subsection{Case Study}
\begin{table*}[h]
\centering
\begin{tabular}{|p{0.95\textwidth}|}
\hline
\textbf{Query:} What is the order and rules for exception handling? \\
\hline
\textbf{Human:}
In Python, the order and rules for exception handling are quite important. Understanding these rules can help us handle potential errors in our programs more effectively.

1. Order Rules:
   - When an exception is raised within a `try` block, Python immediately stops executing the code within that block. \\
   
   - Python then checks the `except` clauses in the order they appear to see if any of them match the raised exception. \\
   - If a match is found, the corresponding `except` block is executed. \\
   - If no matching `except` clause is found, the exception is passed to the outer `try-except` blocks, or if the exception remains unhandled throughout the program, the Python interpreter will print an error message and exit the program. 

2. Matching Rules:
   - Each `except` clause can specify the type of exception it wants to handle. Python first checks if the exception type specified in the first `except` clause matches the type of the raised exception. \\
   - If there is a match, the code in that `except` block is executed, and then the program continues with the remaining code. \\
   - If there is no match, Python proceeds to check the next `except` clause. \\
   - An `except:` clause without a specified exception type will catch all exceptions. \\

3. Exception Type Hierarchy:
   - Exceptions are hierarchical. If an `except` clause catches a base class exception, it will also catch all exceptions that are subclasses of that base class. \\
   - Therefore, when defining `except` clauses, start with the most specific exception types and move to more general ones to avoid catching unexpected exceptions too early. \\

4. Multiple Exceptions Handling:
   - You can follow a single `try` block with multiple `except` clauses to handle different types of exceptions. \\
   - Multiple exceptions can be caught in a single `except` clause by listing the exception types in a tuple. \\

5. `else` and `finally` Blocks:
   - If the `try` block does not raise any exceptions, the `else` block will be executed. \\
   - The `finally` block will execute regardless of whether an exception was raised or not, and is typically used for cleanup actions such as closing files or releasing resources. \\

Here’s an example:
\\
try: 
    \# Code that may raise an exception
    pass
except ValueError:
    \# Handle ValueError
    pass
except (TypeError, ZeroDivisionError):
    \# Handle TypeError or ZeroDivisionError
    pass
except Exception as e:
    \# Handle all other exceptions and access the exception object
    print(f"An error occurred: {e}")
else:
    \# Execute this block if no exceptions were raised
    pass
finally:
    \# Execute this block regardless of whether an exception was raised or not
    pass
```
 \\
\hline
\textbf{Molly:} 
When an error occurs during the execution of a Python script, the Python interpreter creates an exception object. If this exception is not caught and handled, the interpreter will stop the execution of the script and output an error message.

To handle exceptions, you use the `try`...`except` statement. You place the code that might trigger an exception inside the `try` block, and handle any exceptions that occur within the `except` block. The detailed handling process is as follows:

1. Exception Matching: When an exception is triggered inside the `try` block, the Python interpreter checks if this exception matches any of the types specified in the `except` blocks, from top to bottom. Therefore, you typically place more specific exception types before more general ones.

2. Inheritance Relationship: During exception matching, Python uses the relationship between subclasses and parent classes. A subclass exception will match its parent class exception, but not vice versa. For example, a `KeyboardInterrupt` exception will be caught by `except Exception`, but `Exception` will not be caught by `except KeyboardInterrupt`.

3. Handling Multiple Exceptions: If multiple exceptions occur, the Python interpreter will handle only the first triggered exception.

4. No Exception Handling: If the `try` block does not raise any exceptions, the `except` clauses are ignored. If an `else` clause is present, it will be executed.

5. Unmatched Exceptions: If an exception is raised and not matched by any `except` block, it will be passed to the outer `try` block, or if it reaches the top level (script execution level), the script will stop and an error message will be output.

6. Final Cleanup: Regardless of whether an exception was raised or not, if there is a `finally` block, it will always be executed. This is used for cleanup actions, such as closing file streams or releasing resources.

Here is an example to illustrate these concepts:

```python
try:
    \# Code that might raise an exception
    pass
except ValueError:
    \# Handle ValueError
    pass
except (TypeError, ZeroDivisionError):
   \# Handle TypeError or ZeroDivisionError
    pass
except Exception as e:
    \# Handle all other exceptions and access the exception object
    print(f"An error occurred: {e}")
else:
    \# Execute this block if no exception was raised
    pass
finally:
    \# Execute this block regardless of whether an exception was raised or not
    pass
```

This explains the order and rules of exception handling in Python.

In summary, understanding and correctly applying the order and rules of exception handling can make your Python programs more robust, clear, and easier to debug and maintain. It's like following traffic rules to ensure smooth and safe code execution. \\
\hline

\end{tabular}
\caption{Answer of Human and answer generated by Molly (GPT-4).}
\label{tab:c}
\end{table*}

In this section, we provide some cases in Table \ref{tab:c}. Using a same question, we provide the answer of human and answer generated by Molly.

\section{Conclusions}
In this paper, we propose a effective framework for Python programming education called Molly. This framework allows for the gradual construction of target outcomes and performs iterative self-correction. To align the knowledge of large language models with that of human programming experts, we have constructed a Python question-and-answer database for programming education in Chinese. This involves structuring educational materials from Python teaching documents and providing explanations and case studies for relevant questions. This approach enables the large model to learn from human teaching experience, thereby enhancing the usefulness of the generated answers. Additionally, we leverage the role-playing mechanism of large models to simulate the teacher-student questioning process in real-world scenarios. By providing concise answer descriptions, we enrich the context of the learner's questions, which enhances the accuracy of answer retrieval. Finally, the answers are optimized through an iterative self-reflection mechanism, enhancing their usability and usefulness for Python beginners. Experimental results demonstrate that MollyAgent significantly improved the Python teaching capabilities of large language models. 

The dataset proposed in this paper primarily targets beginner Python learners. In the future, we plan to further optimize our structured dataset to enhance its educational value and relevance. Additionally, we will introduce personalized answers to tailor the responses generated by the large language model to different types of learners. We also aim to further refine the proposed intelligent agent framework to achieve better integration and robustness. Additionally, due to the subjectivity of manual evaluation, we will design automated evaluation methods to assess the capabilities of large language models in answering Python programming-related questions.

\bibliography{aaai25}

\end{document}